\documentclass[conference]{IEEEtran}

\usepackage[pdftex]{graphicx}

\ifCLASSINFOpdf{}
\else
\fi
\usepackage{hyperref}

\usepackage{dblfloatfix}

\usepackage[most]{tcolorbox}
\usepackage{lipsum}

\definecolor{linequote}{RGB}{224,215,188}
\definecolor{backquote}{RGB}{245,245,245}

\newtcolorbox{myquote}[1][]{%
    enhanced, breakable, 
    size=minimal,
    frame hidden, boxrule=0pt,
    sharp corners,
    colback=backquote,
    #1
}
\usepackage{subcaption}
\usepackage{listings}
\usepackage{lipsum}

\usepackage[super]{nth}

\usepackage[]{booktabs}

\begin{document}
%
\title{AiBAT\@: Artificial Intelligence/Instructions \\for Build, Assembly, and Test}

\thispagestyle{plain}
\pagestyle{plain}

\author{\IEEEauthorblockN{Benjamin Nuernberger\IEEEauthorrefmark{1},
Anny Liu\IEEEauthorrefmark{2},
Heather Stefanini\IEEEauthorrefmark{3},  Richard Otis\IEEEauthorrefmark{4}, Amanda Towler\IEEEauthorrefmark{5}, R. Peter Dillon\IEEEauthorrefmark{6}
\IEEEauthorblockA{Jet Propulsion Laboratory,
California Institute of Technology,  Pasadena, CA USA\\
Email: \IEEEauthorrefmark{1}benjamin.nuernberger@jpl.nasa.gov,
\IEEEauthorrefmark{2}yaqi.liu@jpl.nasa.gov,
\IEEEauthorrefmark{3}heather.p.stefanini@jpl.nasa.gov,\\
\IEEEauthorrefmark{4}richard.otis@outlook.com, \IEEEauthorrefmark{5}amanda.towler@jpl.nasa.gov, \IEEEauthorrefmark{6}robert.p.dillon@jpl.nasa.gov}}}

\IEEEoverridecommandlockouts
\IEEEpubid{\begin{minipage}{\textwidth}\ \\[12pt] ~\copyright~2024. California Institute of Technology. Government sponsorship \\acknowledged.
\end{minipage}}

\maketitle

\IEEEpubidadjcol

\begin{abstract}
Instructions for Build, Assembly, and Test (IBAT) refers to the process used whenever any operation is conducted on hardware, including tests, assembly, and maintenance. 
Currently, the generation of IBAT documents is time-intensive, as users must manually reference and transfer information from engineering diagrams and parts lists into IBAT instructions. 
With advances in machine learning and computer vision, however, it is possible to have an artificial intelligence (AI) model perform the partial filling of the IBAT template, freeing up engineer time for more highly skilled tasks. 
AiBAT is a novel system for assisting users in authoring IBATs. 
It works by first analyzing assembly drawing documents, extracting information and parsing it, and then filling in IBAT templates with the extracted information. 
Such assisted authoring has potential to save time and reduce cost.
This paper presents an overview of the AiBAT system, including promising preliminary results and discussion on future work.
\end{abstract}


%
\IEEEpeerreviewmaketitle{}

\section{Introduction}\label{sec:introduction}

\begin{figure}[hbt]
\centering
\includegraphics[width=0.9\columnwidth]{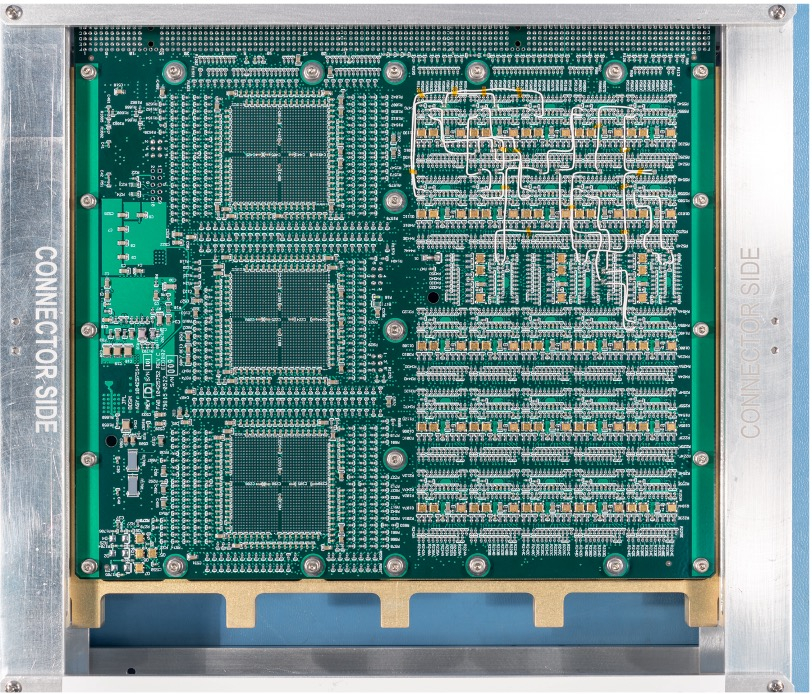}
\caption{A printed wiring assembly (PWA) for the Europa Clipper mission~\cite{phillips2014europa} is built, assembled, and tested using IBAT documents authored from assembly drawings.}\label{fig_pwa}
\end{figure}

At the National Aeronautics and Space Administration (NASA) Jet Propulsion Laboratory (JPL), the IBAT process is used to document how projects are fabricating, building, assembling, and testing their hardware~\cite{ibatbrief,IBATWebsite}. 
The IBAT process has been used extensively on missions such as the Mars Perseverance Rover~\cite{farley2020mars}, the Europa Clipper~\cite{phillips2014europa} mission, and many others.
For example, IBATs are used when building printed wiring assemblies, such as the one shown in Figure~\ref{fig_pwa}.
IBAT documents have three primary functions: 
\begin{enumerate}
    \item providing a way for the user to plan operational steps to be performed on hardware,
    \item providing the instructions to follow during the execution of operations, and
    \item serving as a record of what was done.
\end{enumerate}
Currently, the authoring of IBAT documents is time and labor intensive, as users must manually reference and transfer numerous amounts of information from engineering diagrams and parts lists into the IBAT template. 
With recent advances in AI, however, it is possible to have an AI model perform the partial filling of the IBAT template, freeing up engineer time for more highly skilled tasks.

To evaluate the viability of this approach, we developed a simple proof-of-concept system that automates locating and extracting information from assembly drawing documents and then assists in filling in IBAT templates from the Electronic Fabrication (EFAB) division at JPL\@. 
Due to their proprietary nature, we cannot show any examples of JPL assembly drawings nor IBATs here. 
However, we note that the assembly drawing is a fairly typical drawing that one might find in other organizations, and for the purposes of AiBAT, we note that it has a list of drawing notes on the first page; such notes are the target of our data extraction. 
Regarding IBAT documents, we note that they contain list of steps for a technician to perform or for quality assurance to sign-off on.

The challenges involved in this work include (1) extracting accurate information from the engineering drawings; and (2) utilizing complex natural language understanding to correlate information between diagrams and IBATs. 
Understanding which information from the diagram corresponds to a given section of the IBAT is not straightforward; sometimes there is an exact keyword match, but other times there is not, or there are multiple instances of the same keyword, of which only one is correct in reference to the specific IBAT section. 
This is where the latest Large Language Models (LLMs) show promise, with their impressive ability to understand complex relationships across disparate datasets and retrieve nuanced answers to natural language questions~\cite{achiam2023gpt}. 

There are several reasons why a machine learning approach is needed to connect engineering data to IBAT generation, rather than a simple software engineering approach.
First, as mentioned above, there are language ambiguities and nuances that exist between assembly drawings and IBAT steps; as a result, these cannot be simply automated without enormous workflow changes and complicated software architecture to connect various systems.
Second, since assembly drawing PDFs are considered the ``signed off'' authoritative documents between various organizations, the raw data may not be accessible in a programmatic way (via APIs).
The advantage of the AiBAT approach is that it is inherently designed to handle language nuances, and it supplements the current IBAT process, thus not interfering with existing processes (e.g., it directly utilizes the ``signed off'' documents).

The main contributions of this work are:
\begin{itemize}
    \item A proof-of-concept, end-to-end workflow called AiBAT that automates information extraction from engineering drawings and the filling in of IBAT steps, including preliminary quantitative results. According to our awareness, this is the first use of LLMs for automating the authoring of spacecraft assembly instructions.
    \item A discussion on risk, cost, and ways forward in this area.
\end{itemize}

\begin{figure*}[hbt]
\centering
\includegraphics[width=0.8\textwidth]{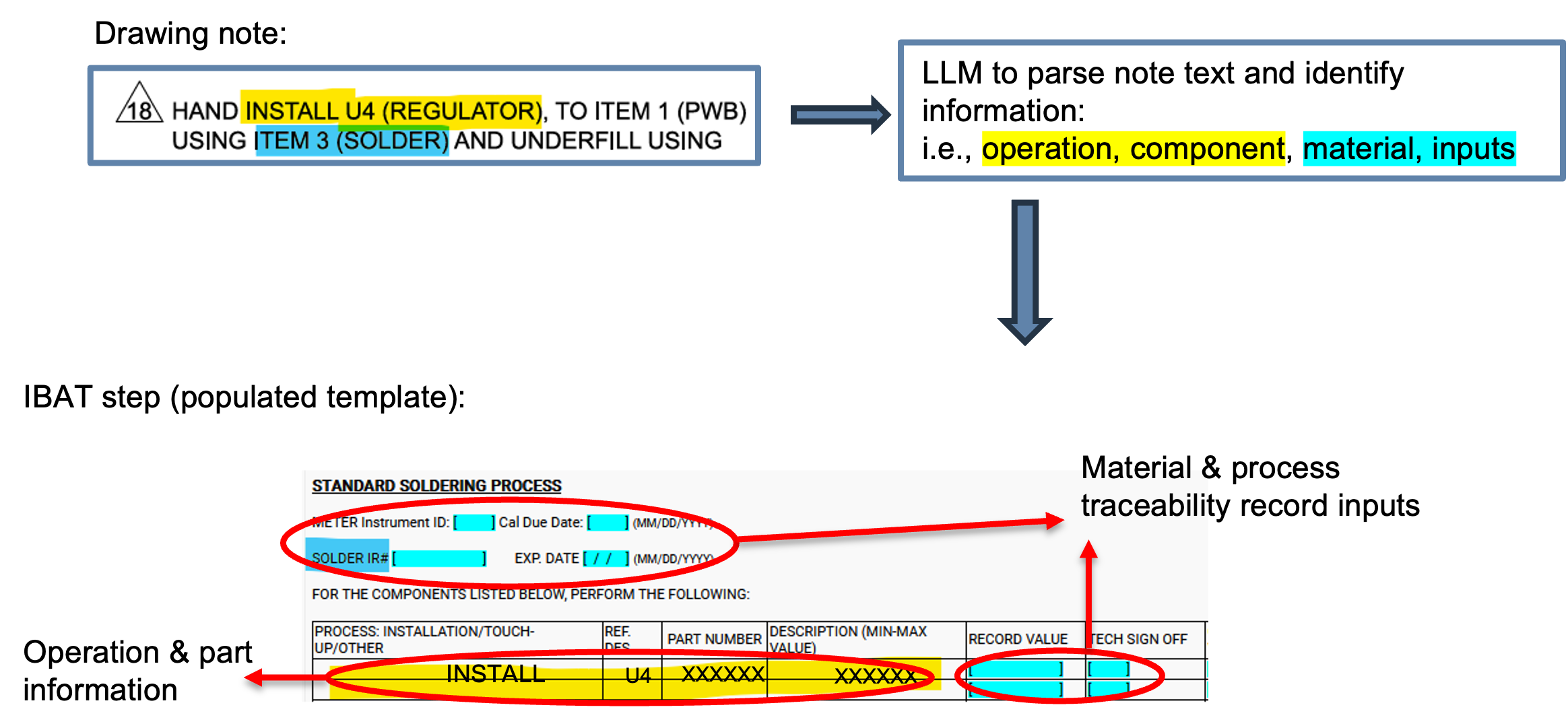}
\caption{The conceptual workflow of AiBAT\@. Here, a drawing note is parsed by the LLM into a set of operations, components, material, and items. The IBAT step template is then populated via the extracted information; here, the action ``INSTALL'' with the reference designator ``U4'' is inserted into the table.}\label{fig_ibat_process}
\end{figure*}

\section{Related Work}
In this section, we described the related work in document understanding, assistive document authoring, and AI/ML for industrial use cases.

\subsection{Document Understanding}
The research field of automatically extracting information from documents via machine automated approaches has previously been referred to as Document Analysis and Recognition~\cite{marinai2008machine}. 
More recently it has been discussed as Document Visual Question Answering (DocVQA)~\cite{mathew2021docvqa}, Visual Document Understanding~\cite{kim2022ocr}, and Visually Rich Document understanding~\cite{xu2020layoutlmv2,ding2023form,ding2024m3}.
On the one hand, the task of document understanding requires raw information extraction via optical character recognition (OCR)~\cite{smith2007overviewtesseractocr}; on the other hand, document understanding may also involve parsing tabular data, understanding figures, charts, and images, etc.
Aballah et al.~\cite{abdallah2024transformers} describe how early methods typically relied on rule-based approaches, while the newest approaches have embraced the Transformer machine learning architecture~\cite{vaswani2017attention}.
Only recently have multimodal LLMs been applied to understanding engineering documentation~\cite{doris2024designqa}.
As detailed in Section~\ref{sec:information_extraction}, we explored a variety of DocVQA models but ultimately relied on a custom rule-based approached for our prototype; future work will likely utilize the latest Transformer-based approaches.

\subsection{Assistive Document Authoring}
Assisting users in authoring documents comes in a variety of form factors.
On the one hand, assistive document authoring may involve a highly structured approach of collecting data and filling in template documents. 
Achachlouei et al.~\cite{achachlouei2021document,achachlouei2023document} present a comprehensive review of document automation techniques, noting the abundance of commercial software in the legal domain for this type of assistive authoring. 
On the other hand, assistive document authoring may also involve a less constrained approach, such as the system offering feedback and guidance or via the user asking the system to write text based on a simple prompt. 
In this regard, OpenAI has showcased the intriguing capabilities of LLMs to write essays to pass simulated exams, or to write descriptive text for images~\cite{achiam2023gpt}. 
Products like Grammarly~\cite{Grammarly} have also gained much traction for generic assistive writing tasks.

The AiBAT system involves a semi-structured approach since the IBAT document (1) utilizes the inherent structure of assembly drawings (e.g., notes and tables and figures) and (2) may have a pre-defined template available for filling in data (as is the case with the EFAB division at JPL). 
In our current prototype implementation, we utilize ``golden IBAT templates'' in our system; however, most IBAT processes do not have templates available and we leave generalizing the capability of the system to support this to future work. 

\subsection{AI/ML for Industrial Use Cases}
There has recently been a large interest in applying AI for industrial use cases, especially in what is known as the \nth{4} Industrial Revolution (or Industry 4.0)~\cite{peres2020industrial}.
More recently, LLMs have been applied throughout the product development lifecycle~\cite{makatura2024can}, including design~\cite{doris2024designqa,gopfert2024opportunities} as well as with conversational assistants~\cite{strano2024hero}.

JPL has recently investigated a variety of Industry 4.0 workflows.
This includes immersive metaverse technologies, such as visualizing CAD models in augmented reality~\cite{berndt2023universe}, immersively visualizing physics data for mission design~\cite{nuernberger2023visualizing}, and using augmented reality for hardware maintenance~\cite{kellogg2023augmented, nuernberger2020under, braly2019augmented}.
We have also begun investigating a variety of use cases for generative AI for JPL, including science and industrial use cases~\cite{wilson2023using, mauceri2023harnessing}.
The AiBAT project can be considered an application of AI to this \nth{4} Industrial Revolution in building spacecraft.

\section{IBAT Authoring and System Overview}\label{sec:ibat_authoring}
IBATs are a ubiquitous part of flight project work at JPL and used across the Lab (e.g., for EFAB, mechanical fabrication, environmental testing, etc.). 
Currently, IBATs are manually written by Subject Matter Experts (SMEs), often involving a repetitive, manual process of copying information from various sources (e.g., drawings, bill-of-materials, etc.) into the IBAT\@. 
For EFAB on Clipper~\cite{phillips2014europa}, it is estimated to take 10--20 hours per IBAT draft –- and there are over 6,000 Clipper IBATs. 
Numerous SMEs report that a large portion of drafting an IBAT is the task of copying information, an ideal task for automation by AI/ML, which could free up SME cognitive load for more challenging tasks. 
The long-term impact is to reduce the time, effort, errors and, ultimately, cost associated with the IBAT creation process.

Figure~\ref{fig_ibat_process} describes the conceptual workflow of how we use LLMs to parse the assembly drawing note information and to insert the relevant information into an IBAT template.
As previously noted, not all workflows utilize IBAT templates; EFAB, however, has templates available, and we utilize these to provide assistive authoring of the IBAT\@.

\begin{figure*}[hbt]
\centering
\includegraphics[width=0.8\textwidth]{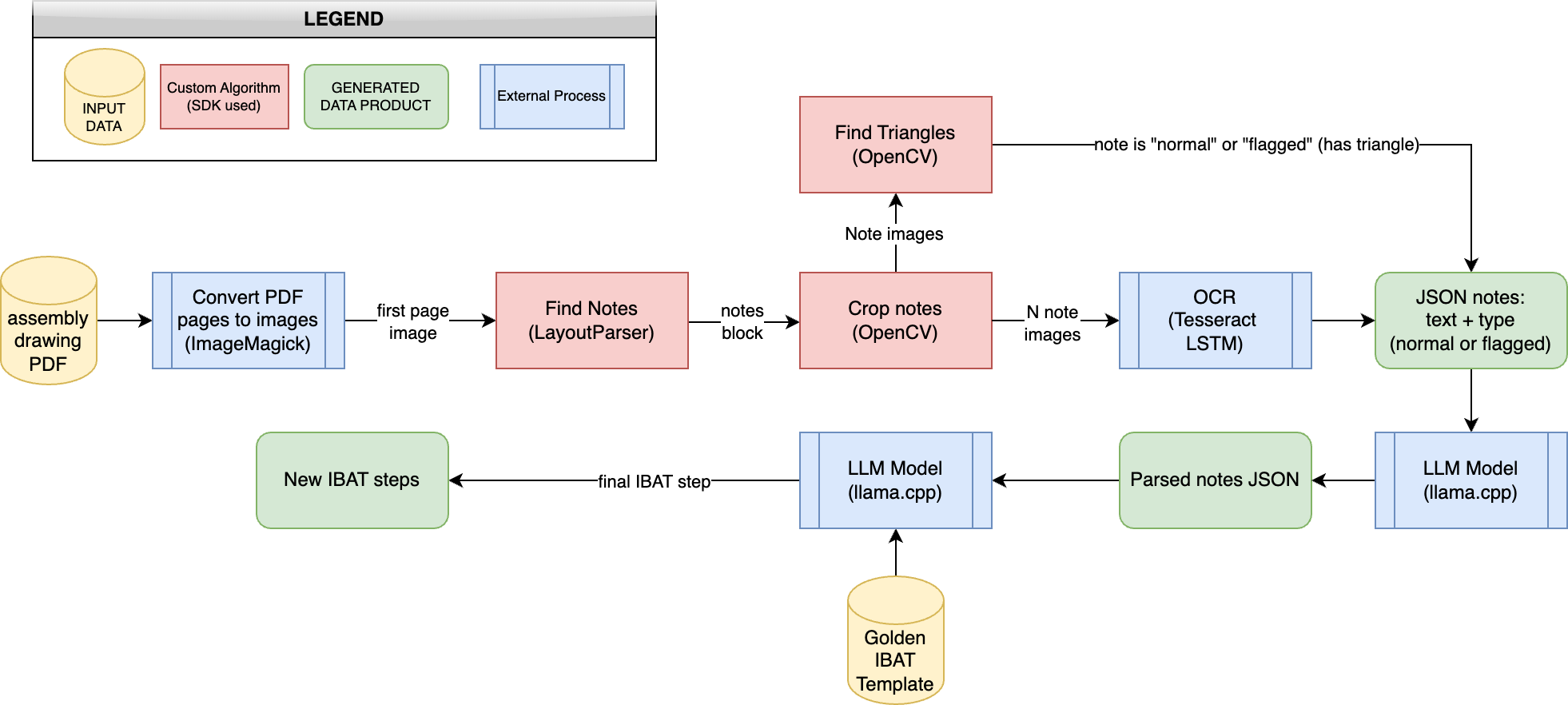}
\caption{AiBAT system architecture diagram. We first use a custom image processing approach to extract the assembly drawing notes (see Section~\ref{sec:information_extraction}). We then call the LLM twice, first to parse the notes into actions, information, and entities, and then to generate the final IBAT steps via using the golden IBAT template steps. }\label{fig_sys_arch}
\end{figure*}

Figure~\ref{fig_sys_arch} provides a high level overview of the AiBAT system.
Section~\ref{sec:information_extraction} goes into the details of how we extract information from the assembly drawings.
Section~\ref{sec:llm} then describes the LLM parsing of notes as well as the final IBAT step generation.

\section{Information Extraction}\label{sec:information_extraction}

\subsection{DocVQA Testing}
Due to export control restrictions, we were unable to test our specific data on the latest multimodal DocVQA models.
However, we conducted a preliminary investigation into how well similar, publicly available assembly documents could be understood by some popular recent DocVQA models~\cite{kim2021donut, liu2024visual, bai2023qwen, achiam2023gpt}.
Smaller and less accurate models failed quickly~\cite{kim2021donut, liu2024visual}; however, the larger and more accurate models showed strong potential for this use case~\cite{bai2023qwen, achiam2023gpt}.
We also found that commercial solutions showed very strong potential~\cite{AzureAIDocIntelligence}.
Due to export control restrictions and limited time, we decided to pursue a simple custom rule-based approach, as described in the next section.

\subsection{Custom Approach}
As noted previously, assembly drawing PDFs are considered the ``signed off'' authoritative documents between various organizations.
While some PDFs may have selectable text that is directly extractable via PDF SDKs, there is no guarantee that a given assembly drawing PDF has that characteristic (in fact, one of our test PDFs fell into this category).
Thus, we opted for the more general case of using OCR to extract the text from the assembly drawings.

\begin{figure}[thb]
    \centering
    \includegraphics[width=0.65\columnwidth]{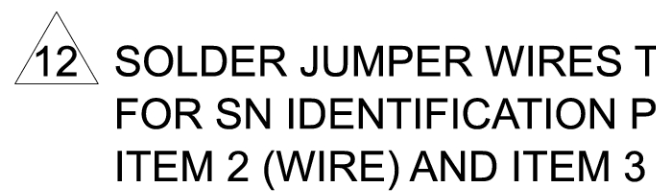}
    \caption{A cropped screenshot of a flagged note. Flagged notes are indicated by the triangle shape around the note number.}\label{fig_flagged_note}
\end{figure}

In our custom approach, we first convert the assembly drawing PDF into a set of images using ImageMagick~\cite{still2006definitiveImageMagick}.
Then, to detect where the assembly notes are, we utilize
LayoutParser~\cite{shen2021layoutparser}, using the
Detectron2 architecture~\cite{wu2019detectron2} and Faster R-CNN Model~\cite{ren2015faster}, trained on the TableBank dataset~\cite{li2020tablebank}.
This effectively provides a cropped image of the assembly notes.
We then apply a simple rule-based image processing algorithm to further crop out individual note images.
We utilize OpenCV~\cite{bradski2000opencv} to detect columns and rows based on simple rules such as ensuring that a certain percentage of consecutive pixels are white.

Assembly drawings sometimes have flagged notes which are notes that have their number surrounded by a triangle shape; see Figure~\ref{fig_flagged_note}.
We utilize two approaches to detect and remove these triangle shapes so that the final OCR can more accurately detect the note number.
First, we utilize a contour detection approach~\cite{suzuki1985topological}, followed by a triangle approximation routine using the Ramer–Douglas–Peucker algorithm. 
This method works well for thick triangles.
To remove the triangle, we simply draw a white triangle on top of any found triangles.
The second approach, which we found to work better for thin triangles, is to perform a dilation and erosion operation.
The dilation will remove any thin lines, while the subsequent erosion will effectively make remaining black pixels (text) be brought back to their original thickness.
If there is a notable image difference between the original image and this processed image, we conclude that there must have been a triangle around the note number.

Finally, OCR is performed on the final individually cropped (and triangle removed) note images using Tesseract~\cite{smith2007overviewtesseractocr}, with the LSTM~\cite{hochreiter1997long} configuration.

\section{LLM Note Parsing and Final Step Generation}\label{sec:llm}
Due to the export controlled nature of most of our data, we opted to utilize on-premise LLMs for our prototyping~\cite{LMStudio,Ollama,llama.cpp}.
We designed our system to be able to utilize various LLM frameworks, but mostly used llama.cpp~\cite{llama.cpp} since it supported JSON schema and was relatively easy to deploy on JPL's High Performance Computing (HPC) cluster.
We tested a variety of models and ended up using Mistral 7B most of the time~\cite{jiang2023mistral7b}.

We explored a variety of prompt techniques, including zero-shot, few-shot, and chain-of-thought~\cite{wei2022chain}.
With initial testing, zero-shot appeared to be too difficult for the LLM to achieve accurate results and chain-of-thought~\cite{wei2022chain} was perhaps too complicated for our initial prototype.
Thus, in the end, we relied on few-shot prompting.

\begin{figure*}[h]
    \centering
    \begin{myquote}[]
    \begin{lstlisting}
    Your task is to take json input and output a parsed version in json.
    
    Common actions include: "SOLDER", "BOND", ...
    
    The following are common Reference Designators:
    PRT# == thermal couple
    C# == capacitor
    ...
    
    INPUT:
        { 
            "note": "REMOVE REF DES LISTED IN TABLE 4. BOND ITEM 8\n(CIP) TO ITEM 1 (PWB) USING ITEM 7 (EC 55/9) ... OPTIMAL WIRE\nROUTING TO BE DETERMINED PER MANUFACTURING,\nPACKAGING OR COGNIZANT ENGINEER DISCRETION.", 
            "type": "flagged" 
        }
    OUTPUT:
        {   
            "steps": [
                { "action": "REMOVE", "text": "REMOVE REF DES LISTED IN TABLE 4." },
                { "action": "BOND", "text": "BOND ITEM 8\n(CIP) TO ITEM 1 (PWB) USING ITEM 7 (EC 55/9)..." },
                ...
            ],
            "information": [ 
                "OPTIMAL WIRE\nROUTING TO BE DETERMINED PER MANUFACTURING,\nPACKAGING OR COGNIZANT ENGINEER DISCRETION." 
            ],
            "entities": [ 
                { "ref": "REF DES LISTED IN TABLE 4", "type": "reference_designator" },
                { "ref": "TABLE 4", "type": "table" },
                { "ref": "ITEM 8\n(CIP)", "type": "item" },
                { "ref": "ITEM 1 (PWB)", "type": "item" },
                { "ref": "ITEM 7 (EC 55/9)", "type": "item" },
                ...
            ]
        }
    ...
    
    \end{lstlisting}
    \end{myquote}
    \caption{An abridged, example few-shot prompt for parsing a drawing note into a list of actions, information, and entities.}\label{fig:parsing_prompt}
    \end{figure*}

Next, we tested a variety of different strategies in terms of going from assembly drawing note to the final IBAT step content.
One approach was to update the entire IBAT step in one-go; this works by including the entire drawing note and the entire IBAT step template in the prompt and asking the LLM to output the entire final IBAT step.
Another approach was to first break down the IBAT step into smaller chunks (e.g., based upon the IBAT workflow described in Section~\ref{sec:ibat_authoring}) and process
each chunk through a series of LLM calls; this works by first pre-processing the IBAT template step into substeps and also providing specific few-shot prompts for each of those substeps.
In the end we decided to go with this latter approach since during testing it appeared that this would be easier to start with and to also quantify its accuracy.

Finally, we decided to first use the LLM to parse the drawing note into a series of actions, information, and entities (described in Section~\ref{sec:note_parsing}); we chose to take this step to understand how well the LLM can parse out information from the notes.
Following this, we use the parsed notes and the golden IBAT template steps to generate the final IBAT steps (described in Section~\ref{sec:final_step_generation}).
In both uses of the LLM, we enforce the LLM response to conform to a JSON schema, as specified via the llama.cpp server API\@.

\subsection{Note Parsing}\label{sec:note_parsing}
In note parsing, we ask the LLM to output the following:
\begin{itemize}
    \item A list of actions, such as ``BOND'' or ``SOLDER''
    \item A list of information, such as statements referring to reference drawings or other documents
    \item A list of entities, such as items, reference designators, tables, etc.
\end{itemize}

An example, abridged few-shot prompt is shown in Figure~\ref{fig:parsing_prompt}.
We first write out some instructions to the LLM, as well as noting common actions and common reference designators.
The few-shot examples then follow afterwards.

\subsection{Final Step Generation}\label{sec:final_step_generation}
For final step generation, we utilize the parsed note and substeps of the IBAT template steps to output the final IBAT steps.
For example, an ``UNDERFILL'' IBAT template step may include three substeps of (1) a text description of the action to perform; (2) a table with reference designator information; and (3) details on the curing process.
An example few-shot prompt is shown in Figure~\ref{fig:prompt_final_step}.
Notice how we include an ``action'' in the IBAT template portion of the prompt to guide the LLM in what type of action to apply to the IBAT template.
In some situations, we also provide a ``guidance'' field which helps guide the LLM to mitigate errors.

\begin{figure*}[]
\centering
\begin{myquote}            
\begin{lstlisting}
Your job is to take INPUT json and OUTPUT the appropriate json. There are 2 main types of actions listed in the ibat_template items that describe what to do: (1) update; and (2) choose. For the "choose" action, select one item from the list of "options" and keep the chosen item verbatim.

INPUT:
{
    "drawing": "123456789",
    "note": {
        "note_number": 10,
        "type": "flagged",
        "parsed_note": { 
            "steps": [
                { "action": "BOND", "text": "BOND TEMP SENSOR PRT1, PRT2 TO ITEM 1\n(PWB) WITH ITEM 5 (THERMALLY CONDUCTIVE\nMATERIAL)." }
                ...
            ],
            "information": [...],
            "entities": [
                { "ref": "TEMP SENSOR PRT1", "type": "reference_designator" },
                { "ref": "PRT2", "type": "reference_designator" },
                ...
            ]
	  }
    },
    "ibat_template": {
        "type": "text",
        "action": "update",
        "data": "BONDING PRTs\n\nPER DRAWING NOTE X: BOND PRT CERAMIC/WHITE SIDE DOWN AT APPROXIMATE LOCATION SHOWN USING ADHESIVE ITEM X. SERIAL NUMBERS SHOULD BE VISIBLE ON THE TOP SIDE.\nNOTE: DO NOT REMOVE PRT LABEL.\n\nNOTE: IF PACKAGING OF PRT IS SERIALIZED AND THE PRT DOES NOT HAVE THE S/N ON IT, INK STAMP S/N ON PRT."
    }
}
OUTPUT:
{
    "type": "text",
    "data": "BONDING PRTs\n\nPER DRAWING 123456789 NOTE 10: BOND PRT1 & PRT2, CERAMIC/WHITE SIDE DOWN AT APPROXIMATE LOCATION SHOWN USING ITEM 5 (THERMALLY CONDUCTIVE MATERIAL). SERIAL NUMBERS SHOULD BE VISIBLE ON THE TOP SIDE.\n\nNOTE: DO NOT REMOVE PRT LABEL.\n\nNOTE: IF PACKAGING OF PRT IS SERIALIZED AND THE PRT DOES NOT HAVE THE S/N ON IT, INK STAMP S/N ON PRT."
}
...
\end{lstlisting}
\end{myquote}
\caption{An abridged, example few-shot prompt for generating the final IBAT steps. }\label{fig:prompt_final_step}
\end{figure*}

\section{Experimental Evaluation}
We tested our system on a set of 3 IBAT and assembly drawing pairs, which all use the same golden IBAT template:
\begin{enumerate}
    \item Pair 1: 22 drawing notes, with 8 IBAT steps that can be automated, divided into 20 total substeps 
    \item Pair 2: 18 drawing notes, with 8 IBAT steps that can be automated, divided into 17 total substeps 
    \item Pair 3: 22 drawing notes, with 7 IBAT steps that can be automated, divided into 18 total substeps 
\end{enumerate}

\subsection{Information Extraction Results}
Our note extraction approach achieved the following Character Error Rates (avg, std):
\begin{itemize}
    \item Pair 1: 0.002577 (0.005249) 
    \item Pair 2: 0.001903 (0.005204) 
    \item Pair 3: 0.006926 (0.021729) 
\end{itemize}
An example failure case is ``FOR U21'' detected as ``FORU21''; with higher image resolution crops, we expect such errors to not be an issue in the future.
The full extraction pipeline takes approximately 80s to complete on a MacBook M2 Max.

Our triangle detection approach achieved the following accuracy results for correctly detecting if a note is a ``flagged'' note or not:
\begin{itemize}
    \item Pair 1: 100\% (22/22) 
    \item Pair 2: 100\% (18/18) 
    \item Pair 3: 95.5\% (21/22) 
\end{itemize}
In the case of Pair 3, there was a false positive triangle detected for the character ``4'' which appeared in the note text (and has a triangle shape in it).

\subsection{LLM Setup and Metrics}\label{sec:llm_setup}
The following setup was used for LLMs in parsing notes and generating final IBAT steps.
We utilized JPL's High Performance Computing (HPC) cluster GPU nodes which had two NVIDIA A100s per node, and used Mistral 7B~\cite{jiang2023mistral7b} for our experimental evaluation.

Accuracy was judged by a mechanical engineer SME, familiar with both the IBAT and assembly drawing documents, who categorized each LLM output (i.e., each parsed note or each substep) into the following result categories:
\begin{enumerate}
    \item R0: no errors
    \item R1: trivial error (e.g., whitespace difference, minor wording difference, etc.)
    \item R2: minor error (e.g., omitted relevant information, added unnecessary information, etc.)
    \item R3: major error (e.g., wrong info, misplaced info, etc.)
\end{enumerate}

Note that, in some cases, there may be multiple results per LLM output (e.g., both R1 and R2 type errors).
In addition, in the reporting of the results, we also note as ``\%R01'' the percentage of parsed notes or substeps (depending on the task) that have only R0 or R1 results, since these result types indicate that no major edits would be needed to the AiBAT output and that the generated result is correct.

Finally, we note that all few-shot prompts used information only from Pair 1, which we label with an asterisk (*) in the results, since any results for this pair should inherently be better due to the prompts including its information.

\subsection{Parsing of Notes Results}
Automated LLM parsing of notes into steps, information, and entities took an average of 90s per assembly drawing.
Table~\ref{table:parsing_results} gives the results.

\begin{table}
    \begin{center}
    \begin{tabular}{ l c c c c c c } 
        \toprule
        & & \multicolumn{5}{c}{Results} \\ \cmidrule(r){3-7}
        Pair    & \# notes & R0 & R1 & R2 & R3 & \%R01 \\ 
        \midrule
        Pair 1* & 22       & 13 & 9 & 3  & 1  & 81.8\% \\ 
        Pair 2  & 18       & 1  & 15 & 12 & 13 & 27.8\% \\ 
        Pair 3  & 22       & 4  & 27 & 19 & 6  & 54.5\% \\ 
        \bottomrule
    \end{tabular}
    \caption{Parsing Results; see Section~\ref{sec:llm_setup} for an explanation of the R\# metrics.}\label{table:parsing_results}
    \end{center}
\end{table}

\subsection{Final IBAT Step Generation Results}\label{sec:final_results}
To evaluate the final step generation, we utilized ground truth parsed notes, so as to analyze the accuracy of this part of our pipeline individually.
Final step generation took an average of 40s per IBAT assembly drawing pair.
Table~\ref{table:final_step_results} gives the results.

\begin{table}
    \begin{center}
    \begin{tabular}{ l c c c c c c } 
        \toprule
        & & \multicolumn{5}{c}{Results} \\ \cmidrule(r){3-7}
        Pair    & \# substeps & R0 & R1 & R2 & R3 & \%R01 \\ 
        \midrule
        Pair 1* & 20          & 13 & 7  & 2  & 0  & 90.0\% \\ 
        Pair 2  & 17          & 9  & 6  & 2  & 2  & 76.5\% \\ 
        Pair 3  & 18          & 11 & 7  & 3  & 3  & 72.2\% \\ 
        \bottomrule
    \end{tabular}
    \caption{Final Step Generation Results; see Section~\ref{sec:final_results} for an explanation of the R\# metrics.}\label{table:final_step_results}
    \end{center}
\end{table}

\section{Discussion}
\subsection{Results}
First, we note that this initial prototype was only tested with three pairs of assembly drawings and IBATs, and that all pairs were taken from EFAB use cases.
On the one hand, this limits the generalizability of the current AiBAT system; on the other hand, the overall workflow and system architecture are usable, making further handling of additional data doable.

For information extraction, our custom image processing approach was very accurate, and we were thus satisfied with it for the current prototype.
However, future approaches could potentially move away from this custom approach to using multimodal foundation models that would likely be more generalizable to various assembly drawings beyond the EFAB ones.

LLM note parsing results were mixed.
While Pair 1 achieved strong \%R01 results, Pair 2's results were poor.
Most R3 results from Pair 2 were due to missing actions, information, or entities.
Possible improvements may come from more accurate LLMs or better few-shot prompting.
In addition, having additional data readily available can strengthen our few-shot prompts to generalize our system further to a variety of IBAT use cases.
Using more sophisticated prompting techniques, including dynamically created prompts via retrieval augmented generation (RAG)~\cite{lewis2020retrieval}, should also improve accuracy and generalizability.
If necessary, fine-tuning of models may be done as well.

Finally, LLM final step generation results showed extremely promising potential, with only a few R3 results and high \%R01 scores.
Ways to improve here are similiar to that of improving note parsing, such as using newer and more accurate LLMs, strengthening few-shot prompts, using more sophisticated prompting techniques, and fine-tuning models as necessary.

\subsection{Risks}
There are several risks with applying AI to assist in authoring IBATs.
This includes the risks of confabulations, cybersecurity issues, and the possibility of insufficient data.

First, the risk of confabulations (more popularly known as hallunications)~\cite{huang2023survey, sui2024confabulation} refer to the risk of the LLM to generate a false but plausible sounding response.
In the context of build, assembly, and testing of spacecraft hardware, incorrect instructions could lead to hardware damage and personnel safety risks.
While the likelihood of IBAT authoring errors is low with the current manual authoring process, the severity level can be high due to the sensitive nature of spacecraft hardware.
Thus, there are already safeguards in place in the existing IBAT process to reduce such risk, including via reviews by the IBAT author's organization as well as by the Quality Assurance (QA) organization.
In this regard, the AiBAT system could still utilize these existing safeguards to migitate the risk of LLM confabulations.
In addition, the time savings introduced by AiBAT would potentially allow for additional review time by SMEs to ensure IBAT correctness.
Finally, future work should investigate how to determine if the AiBAT system is accurate enough for production deployment (e.g., should the \%R01 metric scores be above a certain threshold?).

Another major risk is related to cybersecurity.
If the AiBAT system eventually utilizes external LLM services, there is a risk of unauthorized access to data in-transit to and from the LLM service, as well to data at-rest if it is cached in the LLM service.
To mitigate these risks, researchers have been studying various approaches to either encrypt or sanitize data before sending it to the LLM~\cite{gilad2016cryptonets,kan2023protecting}.

Finally, there is the risk that we may not have sufficient, quality data to deploy AiBAT more broadly.
On the one hand, due to the nature of building custom spacecraft, sensors, and instruments, every project worked on at JPL is very different, which means that assembly drawings and IBATs may be very different.
On the other hand, the low-level tasks remain similar (e.g., soldering) and thus we can still utilize previous assembly drawing notes and IBAT steps in few-shot prompts.
However, every engineer inherently words things differently from other engineers; and in some cases, assembly drawing notes and/or IBAT steps may be poorly (or incorrectly) worded, in which case we would ideally not use those in few-shot prompt examples for our system.
In addition, sometimes IBAT documents get redlined or reworked, and this therefore reduces the amount of available, quality data.
Overall, we currently believe this risk to be low, but we will have to reevaluate in future work by examining more assembly drawing and IBAT document pairs.

\subsection{Cost}
While the current AiBAT prototype runs on-premise, we put together a cost estimation for running this on Microsoft Azure's OpenAI platform\footnote{The cost information contained in this document is of a budgetary and planning nature and is intended for informational purposes only. It does not constitute a commitment on the part of JPL and/or Caltech.}.
Assuming GPT-4o costs from September 2024 (\$0.005 per 1K prompt tokens, \$0.015 per 1K completion tokens)~\cite{AzureOpenAIPricing}, the cost to parse notes is approximately \$0.039 per step (tokens: 7.2K prompt, 200 completion) and the cost to generate final steps is approximately \$0.024 per step (tokens: 4.2K prompt, 200 completion).
Assuming per IBAT/assembly-drawing we parse around 20 notes and convert around 10 steps, the total cost is approximately \$1 per IBAT/assembly drawing pair.
NASA JPL produces around 7,000 IBATs per year, which translates to around \$7,000 per year in AiBAT LLM cost if this service were to be implemented at scale.
On the one hand, the cost could go up due to more sophisticated prompt engineering, longer few-shot prompts, more expensive models, etc.
On the other hand, the cost could go down due to algorithm optimizations, cheaper models, etc.
We currently estimate a potential of \$1.25M saved per year (minus \$7K cost from AiBAT) if we could deploy such IBAT assisted authoring for all 7K IBATs generated per year at JPL\@.

\subsection{Future Work}
We note three main directions for future work.
First, programmatically or dynamically updating few-shot prompts has been shown to improve accuracy and may help generalize AiBAT further~\cite{wu2022self,khattab2023dspy}.
Second, we plan to expand our testing beyond EFAB data to increase the generalizability of our system.
Third, we plan to broaden the information extraction coverage to include tables, bill-of-materials, and figures.
Finally, we note that accurate parsing of assembly drawing notes could open the way for automated authoring of immersive instructions, which has been explored previously for augmented reality~\cite{Mohr_retargeting}.

\section{Conclusion}
We presented AiBAT, a novel system that can extract information from assembly drawing documents, parse that information, and then utilize it to assistively author steps in the Instructions for Build, Assembly, and Test process.
Results show strong potential for accurate assistive authoring of IBAT steps, which can lead to engineer time saved, ultimately freeing up their time for handling more cognitively demanding tasks.

\section*{Acknowledgment}
We thank Rob Royce (JPL) for valuable feedback during the development of AiBAT\@.
This research was carried out at the Jet Propulsion Laboratory,
California Institute of Technology, under a contract with the National
Aeronautics and Space Administration (80NM0018D0004) and funded through the Data Science Working Group.

\bibliographystyle{IEEEtran}
\bibliography{refs} 

\begin{thebibliography}{10}
\providecommand{\url}[1]{#1}
\csname url@samestyle\endcsname
\providecommand{\newblock}{\relax}
\providecommand{\bibinfo}[2]{#2}
\providecommand{\BIBentrySTDinterwordspacing}{\spaceskip=0pt\relax}
\providecommand{\BIBentryALTinterwordstretchfactor}{4}
\providecommand{\BIBentryALTinterwordspacing}{\spaceskip=\fontdimen2\font plus
\BIBentryALTinterwordstretchfactor\fontdimen3\font minus
  \fontdimen4\font\relax}
\providecommand{\BIBforeignlanguage}[2]{{%
\expandafter\ifx\csname l@#1\endcsname\relax
\typeout{** WARNING: IEEEtran.bst: No hyphenation pattern has been}%
\typeout{** loaded for the language `#1'. Using the pattern for}%
\typeout{** the default language instead.}%
\else
\language=\csname l@#1\endcsname
\fi
#2}}
\providecommand{\BIBdecl}{\relax}
\BIBdecl

\bibitem{ibatbrief}
M.~H. Postma, J.~R. Narva, S.~N. Flanagan, A.~Khaja, L.~A. Williams, P.~L.
  Brandon, J.~A. Holt, and J.~Flores, ``{Instructions for Build, Assemble, and
  Test (I-BAT) Technical Design Document (TDD)},'' \emph{NASA Tech Briefs},
  2014.

\bibitem{IBATWebsite}
TechBriefs, ``{Build, Assemble, Test (BAT) Planning and Execution Resources
  Application},''
  \url{https://www.techbriefs.com/component/content/article/22427-npo49452},
  accessed: 2024-08-06.

\bibitem{farley2020mars}
K.~A. Farley, K.~H. Williford, K.~M. Stack, R.~Bhartia, A.~Chen, M.~de~la
  Torre, K.~Hand, Y.~Goreva, C.~D. Herd, R.~Hueso \emph{et~al.}, ``Mars 2020
  mission overview,'' \emph{Space Science Reviews}, vol. 216, pp. 1--41, 2020.

\bibitem{phillips2014europa}
\BIBentryALTinterwordspacing
C.~B. Phillips and R.~T. Pappalardo, ``Europa clipper mission concept:
  Exploring jupiter's ocean moon,'' \emph{Eos, Transactions American
  Geophysical Union}, vol.~95, no.~20, pp. 165--167, 2014. [Online]. Available:
  \url{https://agupubs.onlinelibrary.wiley.com/doi/abs/10.1002/2014EO200002}
\BIBentrySTDinterwordspacing

\bibitem{achiam2023gpt}
J.~Achiam, S.~Adler, S.~Agarwal, L.~Ahmad, I.~Akkaya, F.~L. Aleman, D.~Almeida,
  J.~Altenschmidt, S.~Altman, S.~Anadkat \emph{et~al.}, ``Gpt-4 technical
  report,'' \emph{arXiv preprint arXiv:2303.08774}, 2023.

\bibitem{marinai2008machine}
S.~Marinai, \emph{Machine learning in document analysis and recognition}.\hskip
  1em plus 0.5em minus 0.4em\relax Springer Science \& Business Media, 2008,
  vol.~90.

\bibitem{mathew2021docvqa}
M.~Mathew, D.~Karatzas, and C.~Jawahar, ``Docvqa: A dataset for vqa on document
  images,'' in \emph{Proceedings of the IEEE/CVF winter conference on
  applications of computer vision}, 2021, pp. 2200--2209.

\bibitem{kim2022ocr}
G.~Kim, T.~Hong, M.~Yim, J.~Nam, J.~Park, J.~Yim, W.~Hwang, S.~Yun, D.~Han, and
  S.~Park, ``Ocr-free document understanding transformer,'' in \emph{European
  Conference on Computer Vision}.\hskip 1em plus 0.5em minus 0.4em\relax
  Springer, 2022, pp. 498--517.

\bibitem{xu2020layoutlmv2}
Y.~Xu, Y.~Xu, T.~Lv, L.~Cui, F.~Wei, G.~Wang, Y.~Lu, D.~Florencio, C.~Zhang,
  W.~Che \emph{et~al.}, ``Layoutlmv2: Multi-modal pre-training for
  visually-rich document understanding,'' \emph{arXiv preprint
  arXiv:2012.14740}, 2020.

\bibitem{ding2023form}
Y.~Ding, S.~Long, J.~Huang, K.~Ren, X.~Luo, H.~Chung, and S.~C. Han,
  ``Form-nlu: Dataset for the form natural language understanding,'' in
  \emph{Proceedings of the 46th International ACM SIGIR Conference on Research
  and Development in Information Retrieval}, 2023, pp. 2807--2816.

\bibitem{ding2024m3}
Y.~Ding, L.~Vaiani, C.~Han, J.~Lee, P.~Garza, J.~Poon, and L.~Cagliero,
  ``M3-vrd: Multimodal multi-task multi-teacher visually-rich form document
  understanding,'' \emph{arXiv preprint arXiv:2402.17983}, 2024.

\bibitem{smith2007overviewtesseractocr}
R.~Smith, ``An overview of the tesseract ocr engine,'' in \emph{Ninth
  international conference on document analysis and recognition (ICDAR 2007)},
  vol.~2.\hskip 1em plus 0.5em minus 0.4em\relax IEEE, 2007, pp. 629--633.

\bibitem{abdallah2024transformers}
A.~Abdallah, D.~Eberharter, Z.~Pfister, and A.~Jatowt, ``Transformers and
  language models in form understanding: A comprehensive review of scanned
  document analysis,'' \emph{arXiv preprint arXiv:2403.04080}, 2024.

\bibitem{vaswani2017attention}
A.~Vaswani, N.~Shazeer, N.~Parmar, J.~Uszkoreit, L.~Jones, A.~N. Gomez,
  {\L}.~Kaiser, and I.~Polosukhin, ``Attention is all you need,''
  \emph{Advances in neural information processing systems}, vol.~30, 2017.

\bibitem{doris2024designqa}
A.~C. Doris, D.~Grandi, R.~Tomich, M.~F. Alam, H.~Cheong, and F.~Ahmed,
  ``Designqa: A multimodal benchmark for evaluating large language models'
  understanding of engineering documentation,'' \emph{arXiv preprint
  arXiv:2404.07917}, 2024.

\bibitem{achachlouei2021document}
M.~A. Achachlouei, O.~Patil, T.~Joshi, and V.~N. Nair, ``Document automation
  architectures and technologies: A survey,'' \emph{arXiv preprint
  arXiv:2109.11603}, 2021.

\bibitem{achachlouei2023document}
------, ``Document automation architectures: Updated survey in light of large
  language models,'' \emph{arXiv preprint arXiv:2308.09341}, 2023.

\bibitem{Grammarly}
TechBriefs, ``{Grammarly: Free AI Writing Assistance},''
  \url{https://www.grammarly.com/}, accessed: 2024-09-03.

\bibitem{peres2020industrial}
R.~S. Peres, X.~Jia, J.~Lee, K.~Sun, A.~W. Colombo, and J.~Barata, ``Industrial
  artificial intelligence in industry 4.0-systematic review, challenges and
  outlook,'' \emph{IEEE access}, vol.~8, pp. 220\,121--220\,139, 2020.

\bibitem{makatura2024can}
L.~Makatura, M.~Foshey, B.~Wang, F.~H{\"a}hnlein, P.~Ma, B.~Deng,
  M.~Tjandrasuwita, A.~Spielberg, C.~Owens, P.~Y. Chen \emph{et~al.}, ``How can
  large language models help humans in design and manufacturing? part 2:
  Synthesizing an end-to-end llm-enabled design and manufacturing workflow,''
  \emph{Harvard Data Science Review}, 2024.

\bibitem{gopfert2024opportunities}
J.~G{\"o}pfert, J.~M. Weinand, P.~Kuckertz, and D.~Stolten, ``Opportunities for
  large language models and discourse in engineering design,'' \emph{Energy and
  AI}, p. 100383, 2024.

\bibitem{strano2024hero}
L.~Strano, C.~Bonanno, F.~Ragusa, G.~M. Farinella, and A.~Furnari, ``Hero-gpt:
  Zero-shot conversational assistance in industrial domains exploiting large
  language models.'' in \emph{IMPROVE}, 2024, pp. 74--82.

\bibitem{berndt2023universe}
S.-H. Berndt, W.~Burke, M.~M. Gandara, M.~Kimes, L.~Klyne, C.~Mattmann,
  M.~Milano, J.~Nelson, B.~Nuernberger, M.~Sekiya \emph{et~al.}, ``From
  universe to metaverse: A leap into virtual collaboration at nasa jpl,''
  \emph{IEEE Transactions on Industrial Cyber-Physical Systems}, 2023.

\bibitem{nuernberger2023visualizing}
B.~Nuernberger, C.~Cochrane, J.~Williams, L.~Klyne, A.~Gottscholl, H.~Kraus,
  A.~R. Soriano, P.~S. Narvaez, C.-C.~N. Huang, K.~Dang \emph{et~al.},
  ``Visualizing spacecraft magnetic fields on the web and in vr,'' in
  \emph{Adjunct Proceedings of the 36th Annual ACM Symposium on User Interface
  Software and Technology}, 2023, pp. 1--3.

\bibitem{kellogg2023augmented}
J.~Kellogg, K.~Andrea-Liner, J.~Jennings, S.~Timms, R.~C. Keramidas, J.~Berry,
  T.~DeLaCruz, K.~Bryant, J.~Crawford, K.~Roth \emph{et~al.}, ``Augmented
  reality assisted astronaut operations in space to upgrade the cold atom lab
  instrument,'' in \emph{Quantum Sensing, Imaging, and Precision Metrology},
  vol. 12447.\hskip 1em plus 0.5em minus 0.4em\relax SPIE, 2023, pp. 87--89.

\bibitem{nuernberger2020under}
B.~Nuernberger, R.~Tapella, S.-H. Berndt, S.~Y. Kim, and S.~Samochina, ``Under
  water to outer space: Augmented reality for astronauts and beyond,''
  \emph{IEEE computer graphics and applications}, vol.~40, no.~1, pp. 82--89,
  2020.

\bibitem{braly2019augmented}
A.~M. Braly, B.~Nuernberger, and S.~Y. Kim, ``Augmented reality improves
  procedural work on an international space station science instrument,''
  \emph{Human factors}, vol.~61, no.~6, pp. 866--878, 2019.

\bibitem{wilson2023using}
B.~D. Wilson, A.~Mishra, A.~Yepremyan, H.~Venkataram, R.~Royce, K.~Pak, and
  B.~Neurnberger, ``Using retrieval augmented generation for search and
  question answering on science \& engineering documents,'' in \emph{AGU Fall
  Meeting Abstracts}, vol. 2023, 2023, pp. IN53A--04.

\bibitem{mauceri2023harnessing}
S.~Mauceri, A.~Mishra, R.~M. Mcgranaghan, A.~A. Mahabal, L.~Mandrake, B.~Smith,
  D.~J. Graf, B.~Nuerenberger, A.~R. Yepremyan, B.~Wilson \emph{et~al.},
  ``Harnessing large language models for research institutions: an example
  based on nasa/jpl use-cases,'' \emph{Authorea Preprints}, 2023.

\bibitem{kim2021donut}
G.~Kim, T.~Hong, M.~Yim, J.~Park, J.~Yim, W.~Hwang, S.~Yun, D.~Han, and
  S.~Park, ``Donut: Document understanding transformer without ocr,''
  \emph{arXiv preprint arXiv:2111.15664}, vol.~7, no.~15, p.~2, 2021.

\bibitem{liu2024visual}
H.~Liu, C.~Li, Q.~Wu, and Y.~J. Lee, ``Visual instruction tuning,''
  \emph{Advances in neural information processing systems}, vol.~36, 2024.

\bibitem{bai2023qwen}
J.~Bai, S.~Bai, S.~Yang, S.~Wang, S.~Tan, P.~Wang, J.~Lin, C.~Zhou, and
  J.~Zhou, ``Qwen-vl: A frontier large vision-language model with versatile
  abilities,'' \emph{arXiv preprint arXiv:2308.12966}, 2023.

\bibitem{AzureAIDocIntelligence}
Azure, ``{Azure AI Document Intelligence},''
  \url{https://azure.microsoft.com/en-us/products/ai-services/ai-document-intelligence},
  accessed: 2024-08-06.

\bibitem{still2006definitiveImageMagick}
M.~Still, \emph{The definitive guide to ImageMagick}.\hskip 1em plus 0.5em
  minus 0.4em\relax Apress, 2006.

\bibitem{shen2021layoutparser}
Z.~Shen, R.~Zhang, M.~Dell, B.~C.~G. Lee, J.~Carlson, and W.~Li,
  ``Layoutparser: A unified toolkit for deep learning based document image
  analysis,'' in \emph{Document Analysis and Recognition--ICDAR 2021: 16th
  International Conference, Lausanne, Switzerland, September 5--10, 2021,
  Proceedings, Part I 16}.\hskip 1em plus 0.5em minus 0.4em\relax Springer,
  2021, pp. 131--146.

\bibitem{wu2019detectron2}
Y.~Wu, A.~Kirillov, F.~Massa, W.-Y. Lo, and R.~Girshick, ``Detectron2,''
  \url{https://github.com/facebookresearch/detectron2}, 2019.

\bibitem{ren2015faster}
S.~Ren, K.~He, R.~Girshick, and J.~Sun, ``Faster r-cnn: Towards real-time
  object detection with region proposal networks,'' \emph{Advances in neural
  information processing systems}, vol.~28, 2015.

\bibitem{li2020tablebank}
M.~Li, L.~Cui, S.~Huang, F.~Wei, M.~Zhou, and Z.~Li, ``Tablebank: Table
  benchmark for image-based table detection and recognition,'' in
  \emph{Proceedings of the Twelfth Language Resources and Evaluation
  Conference}, 2020, pp. 1918--1925.

\bibitem{bradski2000opencv}
G.~Bradski, ``The opencv library.'' \emph{Dr. Dobb's Journal: Software Tools
  for the Professional Programmer}, vol.~25, no.~11, pp. 120--123, 2000.

\bibitem{suzuki1985topological}
S.~Suzuki \emph{et~al.}, ``Topological structural analysis of digitized binary
  images by border following,'' \emph{Computer vision, graphics, and image
  processing}, vol.~30, no.~1, pp. 32--46, 1985.

\bibitem{hochreiter1997long}
S.~Hochreiter and J.~Schmidhuber, ``Long short-term memory,'' \emph{Neural
  computation}, vol.~9, no.~8, pp. 1735--1780, 1997.

\bibitem{LMStudio}
LMStudio, ``{LM Studio},'' \url{https://lmstudio.ai/}, accessed: 2024-08-06.

\bibitem{Ollama}
Ollama, ``{Ollama},'' \url{https://ollama.com/}, accessed: 2024-08-06.

\bibitem{llama.cpp}
llama.cpp, ``{llama.cpp},'' \url{https://github.com/ggerganov/llama.cpp},
  accessed: 2024-08-06.

\bibitem{jiang2023mistral7b}
\BIBentryALTinterwordspacing
A.~Q. Jiang, A.~Sablayrolles, A.~Mensch, C.~Bamford, D.~S. Chaplot, D.~de~las
  Casas, F.~Bressand, G.~Lengyel, G.~Lample, L.~Saulnier, L.~R. Lavaud, M.-A.
  Lachaux, P.~Stock, T.~L. Scao, T.~Lavril, T.~Wang, T.~Lacroix, and W.~E.
  Sayed, ``Mistral 7b,'' 2023. [Online]. Available:
  \url{https://arxiv.org/abs/2310.06825}
\BIBentrySTDinterwordspacing

\bibitem{wei2022chain}
J.~Wei, X.~Wang, D.~Schuurmans, M.~Bosma, F.~Xia, E.~Chi, Q.~V. Le, D.~Zhou
  \emph{et~al.}, ``Chain-of-thought prompting elicits reasoning in large
  language models,'' \emph{Advances in neural information processing systems},
  vol.~35, pp. 24\,824--24\,837, 2022.

\bibitem{lewis2020retrieval}
P.~Lewis, E.~Perez, A.~Piktus, F.~Petroni, V.~Karpukhin, N.~Goyal,
  H.~K{\"u}ttler, M.~Lewis, W.-t. Yih, T.~Rockt{\"a}schel \emph{et~al.},
  ``Retrieval-augmented generation for knowledge-intensive nlp tasks,''
  \emph{Advances in Neural Information Processing Systems}, vol.~33, pp.
  9459--9474, 2020.

\bibitem{huang2023survey}
L.~Huang, W.~Yu, W.~Ma, W.~Zhong, Z.~Feng, H.~Wang, Q.~Chen, W.~Peng, X.~Feng,
  B.~Qin \emph{et~al.}, ``A survey on hallucination in large language models:
  Principles, taxonomy, challenges, and open questions,'' \emph{arXiv preprint
  arXiv:2311.05232}, 2023.

\bibitem{sui2024confabulation}
P.~Sui, E.~Duede, S.~Wu, and R.~J. So, ``Confabulation: The surprising value of
  large language model hallucinations,'' \emph{arXiv preprint
  arXiv:2406.04175}, 2024.

\bibitem{gilad2016cryptonets}
R.~Gilad-Bachrach, N.~Dowlin, K.~Laine, K.~Lauter, M.~Naehrig, and J.~Wernsing,
  ``Cryptonets: Applying neural networks to encrypted data with high throughput
  and accuracy,'' in \emph{International conference on machine learning}.\hskip
  1em plus 0.5em minus 0.4em\relax PMLR, 2016, pp. 201--210.

\bibitem{kan2023protecting}
Z.~Kan, L.~Qiao, H.~Yu, L.~Peng, Y.~Gao, and D.~Li, ``Protecting user privacy
  in remote conversational systems: A privacy-preserving framework based on
  text sanitization,'' \emph{arXiv preprint arXiv:2306.08223}, 2023.

\bibitem{AzureOpenAIPricing}
Azure, ``{Azure OpenAI Service - Pricing | Microsoft Azure},''
  \url{https://azure.microsoft.com/en-us/pricing/details/cognitive-services/openai-service/#pricing},
  accessed: 2024-09-26.

\bibitem{wu2022self}
Z.~Wu, Y.~Wang, J.~Ye, and L.~Kong, ``Self-adaptive in-context learning: An
  information compression perspective for in-context example selection and
  ordering,'' \emph{arXiv preprint arXiv:2212.10375}, 2022.

\bibitem{khattab2023dspy}
O.~Khattab, A.~Singhvi, P.~Maheshwari, Z.~Zhang, K.~Santhanam, S.~Vardhamanan,
  S.~Haq, A.~Sharma, T.~T. Joshi, H.~Moazam \emph{et~al.}, ``Dspy: Compiling
  declarative language model calls into self-improving pipelines,'' \emph{arXiv
  preprint arXiv:2310.03714}, 2023.

\bibitem{Mohr_retargeting}
\BIBentryALTinterwordspacing
P.~Mohr, B.~Kerbl, M.~Donoser, D.~Schmalstieg, and D.~Kalkofen, ``Retargeting
  technical documentation to augmented reality,'' in \emph{Proceedings of the
  33rd Annual ACM Conference on Human Factors in Computing Systems}, ser. CHI
  '15.\hskip 1em plus 0.5em minus 0.4em\relax New York, NY, USA: Association
  for Computing Machinery, 2015, p. 3337–3346. [Online]. Available:
  \url{https://doi.org/10.1145/2702123.2702490}
\BIBentrySTDinterwordspacing

\end{thebibliography}

\end{document}